\documentclass[10pt,twocolumn,letterpaper]{article}

\usepackage{iccv}
\usepackage{times}
\usepackage{epsfig}
\usepackage{graphicx}
\usepackage{amsmath}
\usepackage{amssymb}
\usepackage{multirow}
\usepackage{pifont}
\usepackage{booktabs}
\usepackage{epsfig}
\usepackage{booktabs,multirow}
\usepackage{graphics}
\usepackage{threeparttable}
\usepackage{color}
\usepackage[normalem]{ulem}
\usepackage{multirow}
\usepackage{float}
\usepackage{amsfonts}
\usepackage{bm}
\usepackage{enumitem}
\usepackage[numbers]{natbib}
\usepackage{array}
\usepackage[table]{xcolor}
\usepackage{colortbl}
\definecolor{myy}{RGB}{126,95,0}
\definecolor{mygray}{gray}{.9}
\definecolor{bblue}{RGB}{30,80,120}
\definecolor{mygray1}{gray}{.7}
\usepackage{bm}
\usepackage{mathtools}
\usepackage{subcaption}
\usepackage[nopar]{lipsum}
\usepackage[export]{adjustbox}
\usepackage{bbold}

\definecolor{mygray}{RGB}{127,127,127}
\definecolor{mygreen}{RGB}{93,174,86}
\definecolor{myyellow}{RGB}{1.0, 0.75, 0.0}

\makeatletter
\newcommand{\thickhline}{%
	\noalign {\ifnum 0=`}\fi \hrule height 1pt
	\futurelet \reserved@a \@xhline
}
\makeatother

\newcommand\blfootnote[1]{%
  \begingroup
  \renewcommand\thefootnote{}\footnote{#1}%
  \addtocounter{footnote}{-1}%
  \endgroup
}

\usepackage[breaklinks=true,bookmarks=false]{hyperref}

\iccvfinalcopy 

\def\iccvPaperID{2704} 

\ificcvfinal\fi

\begin{document}

\title{OnlineRefer: A Simple Online Baseline for Referring Video Object Segmentation}

\author{
Dongming Wu$^{1\ddagger}$,
Tiancai Wang$^2$,
Yuang Zhang$^3$,
Xiangyu Zhang$^{2,4}$,
Jianbing Shen$^{5\dagger}$\\
$^1$ Beijing Institute of Technology,
$^2$ MEGVII Technology, 
$^3$ Shanghai Jiao Tong University,\\
$^4$ Beijing Academy of Artificial Intelligence,
$^5$ SKL-IOTSC, CIS, University of Macau \\
{\tt\small wudongming97@gmail.com,}
{\tt\small wangtiancai@megvii.com,}
{\tt\small shenjiangbingcg@gmail.com}
}

\maketitle
\ificcvfinal\thispagestyle{empty}\fi

\begin{abstract}

Referring video object segmentation (RVOS) aims at segmenting an object in a video following human instruction. Current state-of-the-art methods fall into an offline pattern, in which each clip independently interacts with text embedding for cross-modal understanding. They usually present that the offline pattern is necessary for RVOS, yet model limited temporal association within each clip. In this work, we break up the previous offline belief and propose a simple yet effective online model using explicit query propagation, named OnlineRefer. Specifically, our approach leverages target cues that gather semantic information and position prior to improve the accuracy and ease of referring predictions for the current frame. Furthermore, we generalize our online model into a semi-online framework to be compatible with video-based backbones. To show the effectiveness of our method, we evaluate it on four benchmarks, \ie, Refer-Youtube-VOS, Refer-DAVIS$_{17}$, A2D-Sentences, and JHMDB-Sentences.  Without bells and whistles, our OnlineRefer with a Swin-L backbone achieves\textbf{ 63.5 J\&F} and \textbf{64.8 J\&F} on Refer-Youtube-VOS and Refer-DAVIS$_{17}$, outperforming all other offline methods. Our code is available at \href{https://github.com/wudongming97/OnlineRefer}{https://github.com/wudongming97/OnlineRefer}.
\blfootnote{$\dagger$Corresponding author: \textit{Jianbing Shen}. This work was supported in part by the FDCT grants 0154/2022/A3 and SKL-IOTSC(UM)-2021-2023,
the MYRG-CRG2022-00013-IOTSC-ICI grant and the SRG2022-00023-IOTSC grant.
$\ddagger$The work is done during the internship at MEGVII Technology.
}
 
\end{abstract}

\section{Introduction}

\begin{figure}
	\includegraphics[width=\linewidth]{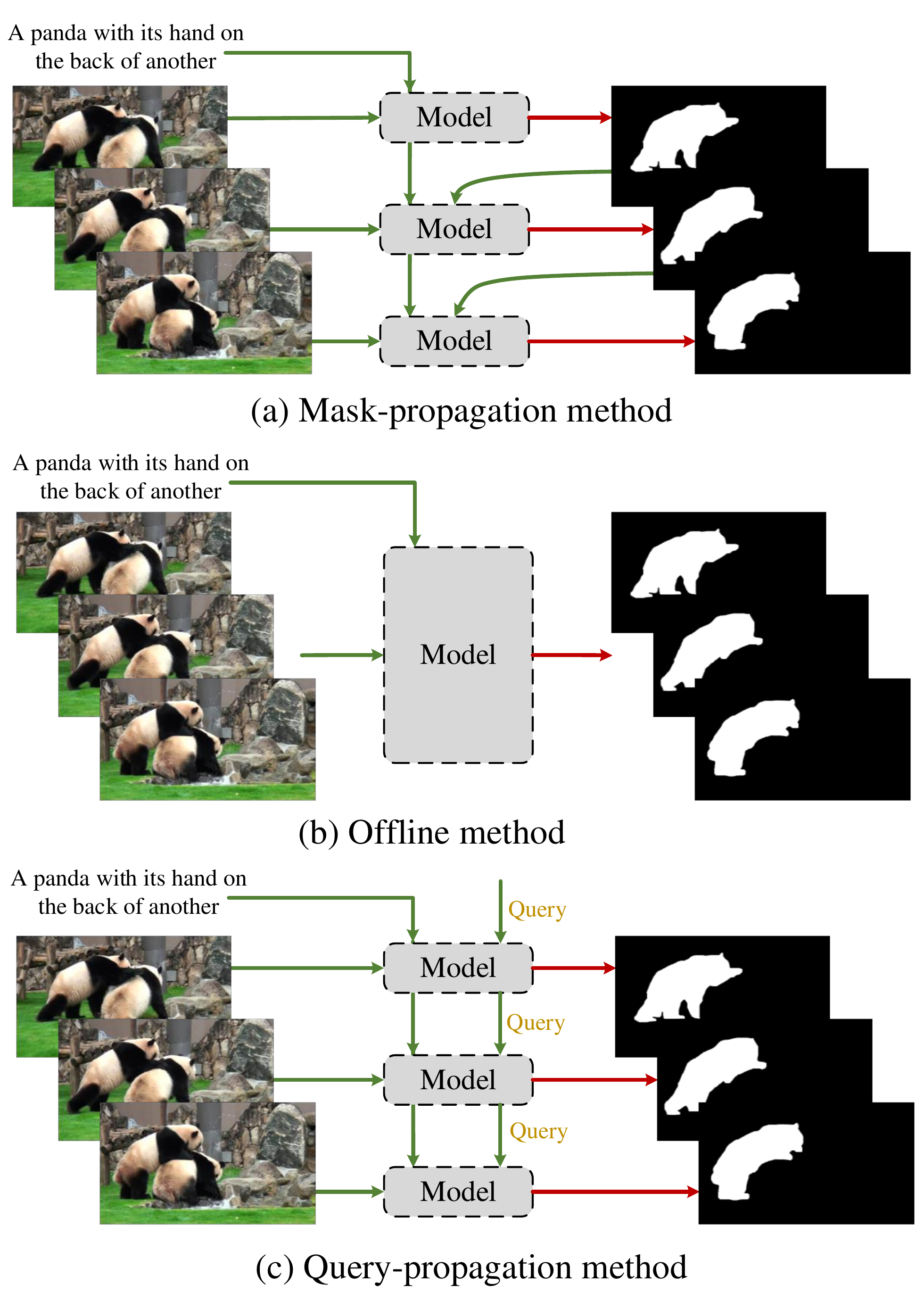}
	\vspace{-20pt}
	\caption{\textbf{Conceptual comparison on current methods}: (a) the  mask-propagation method~\cite{khoreva2019video,URVOS}, (b) the offline method~\cite{hui2021collaborative,mttr,referformer}, and (c) our query-propagation method. }
	\vspace{-4mm}
	\label{fig:motivation}
\end{figure}

Given a natural language expression, the purpose of referring video object segmentation (RVOS) is to segment the described object in a streaming video.
The emerging task has attracted great attention in the computer vision community as it provides potential benefits for many applications, \eg, video editing and human-computer interaction. 
Its core challenge is associating all frames with constructing an efficient video representation, further promoting cross-modal understanding of two modalities, \ie, video and language.
Pioneer methods~\cite{khoreva2019video, URVOS} integrate mask propagation into the referring image segmentation in an online manner, as shown in Fig.~\ref{fig:motivation} (a).
However, the complexity and performance of their model remain far from satisfactory.

\begin{figure*}
\centering
	\includegraphics[width=\linewidth]{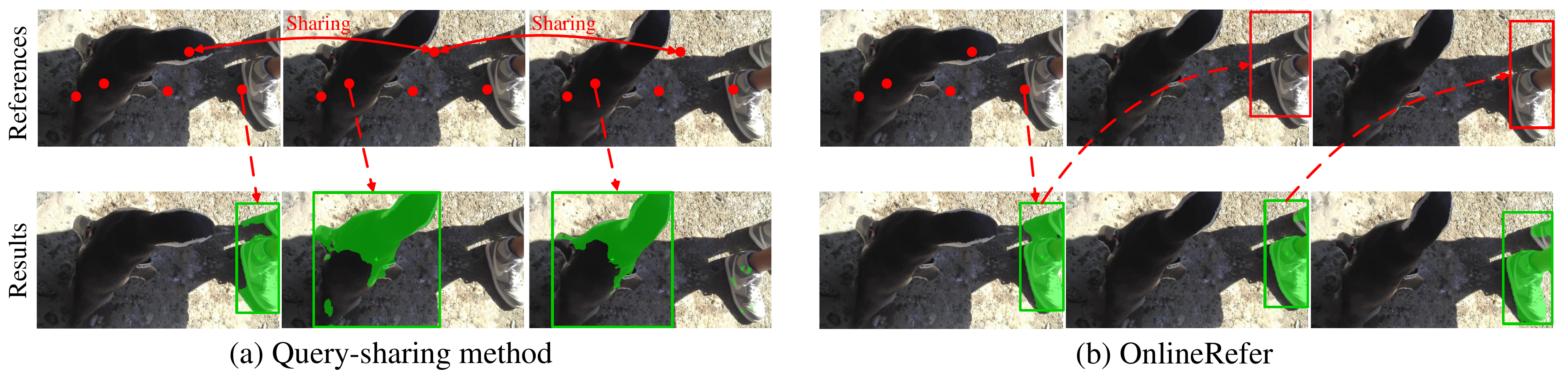}
	\vspace{-24pt}
	\caption{\textbf{Visualization of query references and corresponding results} of (a) query-sharing method and (b) our OnlineRefer. The reference points/boxes are marked \textcolor{red}{red}, while the final predictions of mask and box are marked \textcolor{green}{green}.}
	\vspace{-4mm}
	\label{fig:referpoint}
\end{figure*}

Recently, the state-of-the-art performance on RVOS has been dominated by offline methods~\cite{wang2019asymmetric,hui2021collaborative,mttr,ding2022language,li2022you,referformer,zhao2022modeling}.
They typically follow a clip-level paradigm, dividing the entire video into multiple non-overlapped clips and generating referring object masks for each clip, as illustrated in Fig.~\ref{fig:motivation} (b).
In terms of different inter-frame interaction ways within an individual clip, existing offline methods~\cite{wang2019asymmetric,hui2021collaborative,mttr,ding2022language,li2022you,referformer,zhao2022modeling} can be categorized into two groups: feature association methods and query-sharing methods.
The former feature association methods~\cite{wang2019asymmetric,hui2021collaborative,li2022you,mlsa,zhao2022modeling} integrate multi-frame features into a holistic clip-level visual representation, which is further fused with text embedding for referring prediction.
However, their temporal feature modeling is commonly complicated and heavy-weighted.

In contrast to the feature association methods, the query-sharing methods~\cite{mttr,referformer} provide a simplified pipeline as they build on the query-based Transformer method~\cite{detr,deformable}.
They first construct clip-level cross-modal features and then use a set of repeated queries to retrieve the same referent object from different frames.
In other words, the cross-frame object correspondence relies heavily on sharing input queries.
The interaction between frames is typically limited, hindering the association potential of the learned queries. Fig.~\ref{fig:referpoint} (a) shows a typical example: \textit{all video frames share the same reference points (or queries)},  which misses the occluded object.
In addition, due to resource limitations, the referring prediction has to be performed separately on each clip, which lacks inter-clip association.


In this paper, we propose a new and insightful online referring video object segmentation framework, OnlineRefer.
It goes beyond the intuition of the online model not working well in RVOS.
Its core idea is to take advantage of the query-based set prediction in Deformable DETR~\cite{deformable} and 
link all video frames via continuous query propagation.
Specifically, we first provide a powerful query-based referring segmentation pipeline, which outputs the embedding representations of the referent object, further generating mask, box, and category.
As these outputs gather rich target information, we propose a cross-frame query propagation module to transform them as new query inputs of the next frame.
The propagation process has three significant advantages. 
\textbf{First}, the referring target is automatically associated with its precursors on all previous frames.
\textbf{Second}, the box information of the last frame provides a very good spatial regional prior, benefiting the model for accurately inferring the same object in the current frame (see an example in Fig.~\ref{fig:referpoint} (b)).
\textbf{Third}, our architecture avoids complicated temporal modeling or limited cross-frame association so that the overall training and inference progress is smooth and effective.
Thanks to the remarkable performance, we expect to contribute the elegant and effective online model as a new baseline to the community.

To summarize, our main contributions are three-fold:
\begin{itemize}
  \item We are the first to challenge the widespread belief that only offline models can deal well with RVOS and make online RVOS great again.
  \item We propose a simple yet solid online baseline based on query propagation. The explicit association across video frames facilities temporal target matching and improves referring prediction accuracy.
\item Our method is evaluated on four benchmarks: Refer-Youtube-VOS, Refer-DAVIS$_{17}$, A2D-Sentences,  and JHMDB-Sentences, outperforming all previous offline methods and achieving state-of-the-art performance. 
\end{itemize}

\section{Related Work}

\begin{figure*}
\centering
	\includegraphics[width=0.95\linewidth]{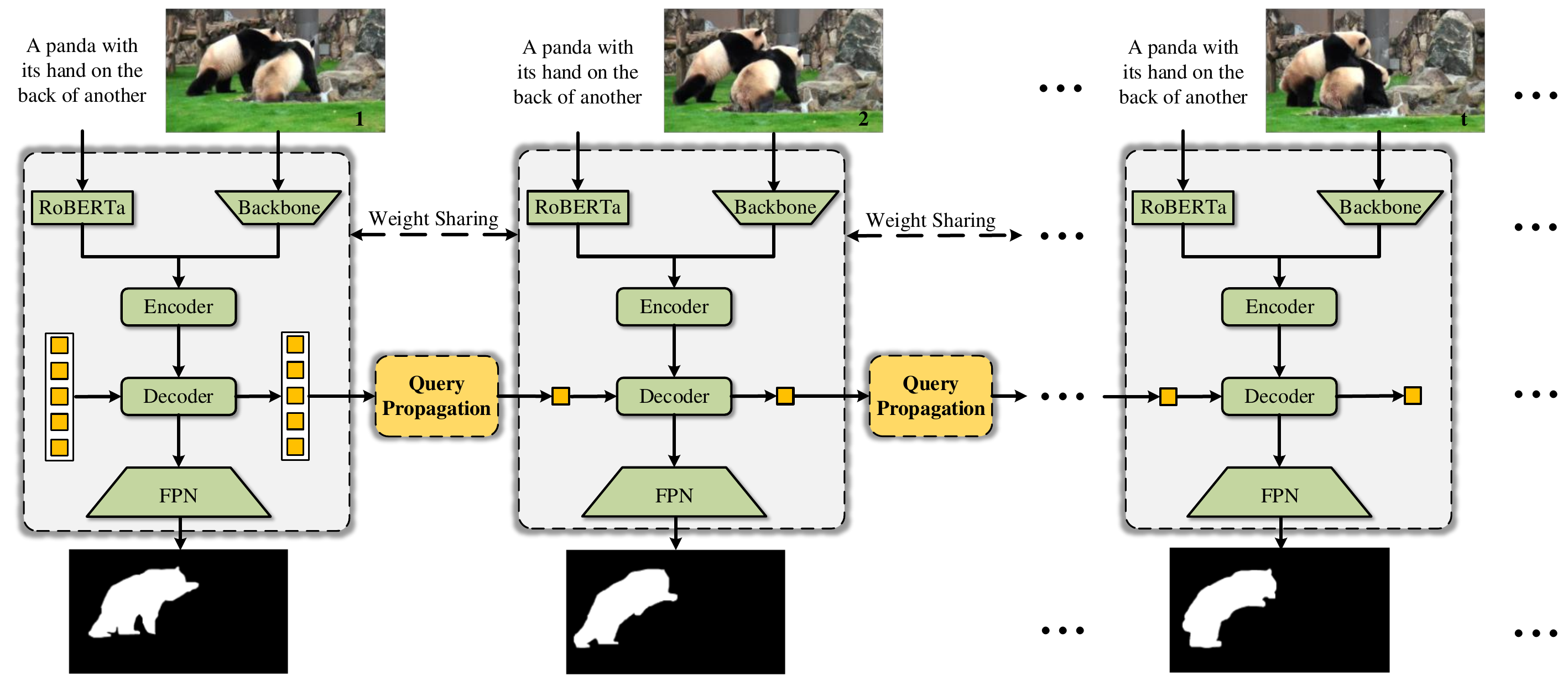}
	\vspace{-6pt}
	\caption{\textbf{Overall architecture of our OnlineRefer.} It consists of two main components: query-based referring segmentation and cross-frame query propagation. The query-based referring segmentation  (\S~\ref{sec:query-based referring segmentation}) employs a set of queries to predict the referring object. The outputs contain rich target-aware information, so the cross-frame query propagation (\S~\ref{sec:query propagation}) transforms them into a new query of the next frame. By repeating the two parts, we complete the RVOS task online frame-by-frame. OnlineRefer can also be generalized into a semi-online baseline using a clip-by-clip manner (\S~\ref{sec:extension}).
 }
	\vspace{-3mm}
	\label{fig:method}
\end{figure*}

\subsection{Referring Video Object Segmentation}

Referring video object segmentation is to localize a text-referred object using a mask.
Earlier works~\cite{khoreva2019video, URVOS} used the spatial-temporal memory mechanism~\cite{stm}, which stored the mask of the previous frames to promote referring image segmentation~\cite{yu2016modeling,CMSA,liu2017recurrent}. 
Currently, however, most existing methods~\cite{wang2019asymmetric,wang2020context,yang2021hierarchical,hui2021collaborative,ding2022language,li2022you,zhao2022modeling,mlsa,yan2023referred} concentrated on designing offline frameworks, \ie, clip-in and clip-out.
For example, Hui~\etal~\cite{hui2021collaborative} proposed a two-stream network, one branch being a temporal encoder to recognize the object motion and another branch being a spatial encoder to generate accurate referring segmentation.
Wu~\etal~\cite{mlsa} additionally considered an object-level branch with salient object regions to enhance the foreground and background discriminability of the model.
These methods inevitably need a complicated spatial-temporal modeling module, which is not trivial.
Several parallel works~\cite{mttr,referformer} employed simple query-based Transformer models, which share the same query set across different frames.
Precisely, MTTR~\cite{mttr} followed an instance-level segmentation pipeline to predict sequences of all instances and select the sequence that best fits the referent object. 
ReferFormer~\cite{referformer} transformed the input expression as the decoder queries for directly attending to the most relevant regions in the video frames.

Although our approach employs the basic query-based prediction architecture, there are two key differences. 
\textit{First}, our model is entirely online and has an inherent advantage in handling long or ongoing videos, while offline methods fail due to the limitation of computation sources.
\textit{Second}, the target information of previous frames is explicitly and effectively used to strengthen cross-frame tracking and frame-wise referring segmentation, achieving better results.

\subsection{Query-based Online Models}
Employing query to associate cross-frame objects has been recently explored in several online models, such as TrackFormer~\cite{trackformer}, MOTR~\cite{motr}, IDOL~\cite{IDOL}, and InsPro~\cite{inspro}.
They show the effectiveness and potential of query-based object association.
However, they are different from our OnlineRefer from two perspectives.
\textit{On the one hand}, OnlineRefer does not need to detect multiple objects due to the guidance of language expression.
TrackFormer~\cite{trackformer} and MOTR~\cite{motr,wu2023referring} adopted an extra query subset to detect new-born objects. They require additional heuristic rules to combine two types of queries \ie,  \textit{track query} and \textit{detect query}.
IDOL~\cite{IDOL} designed a re-identification module as post-processing to link instances between frames.
InsPro~\cite{inspro} kept the fixed query number and employed \textit{empty query}  to detect the new-born objects.
These query propagation methods are also evaluated in \S~\ref{sec:ablation}, while our method performs better.
\textit{On the other hand}, IDOL~\cite{IDOL} and InsPro~\cite{inspro} designed additional training strategies with contrast learning to avoid identification switches or suppress duplicates.
In contrast, our framework minimizes the gap between training and inference of long videos because it avoids the heuristic rules during the training stage.

\section{Methodology}

Given an input video and a natural language expression, our method aims to output binary masks of the referred object in a streaming way.
The overall architecture of OnlineRefer is illustrated in Fig.~\ref{fig:method}. 
It comprises two essential parts: query-based referring segmentation and cross-frame query propagation.
The query-based referring segmentation in \S~\ref{sec:query-based referring segmentation} is an advanced referring segmentation pipeline conditioned on the query set.
The cross-frame query propagation in \S~\ref{sec:query propagation} is to generate the input query set of the current frame by updating the outputs from the last frame.
In addition, for training and inference on video-based backbones, OnlineRefer is extended into a semi-online pattern, which propagates the query across video clips in \S~\ref{sec:extension}.

\subsection{Query-based Referring  Sementation}
\label{sec:query-based referring segmentation}


Similar to ReferFormer~\cite{referformer}, our query-based referring segmentation mainly follows the Deformable DETR detector~\cite{deformable}, and we make several modifications on it for referring object prediction.
It accepts a video frame, a language expression, and a set of learnable queries as input. Its outputs are the target box, mask, category, and a set of output embeddings corresponding to the expression.

\begin{figure}
\centering
	\includegraphics[width=0.9\linewidth]{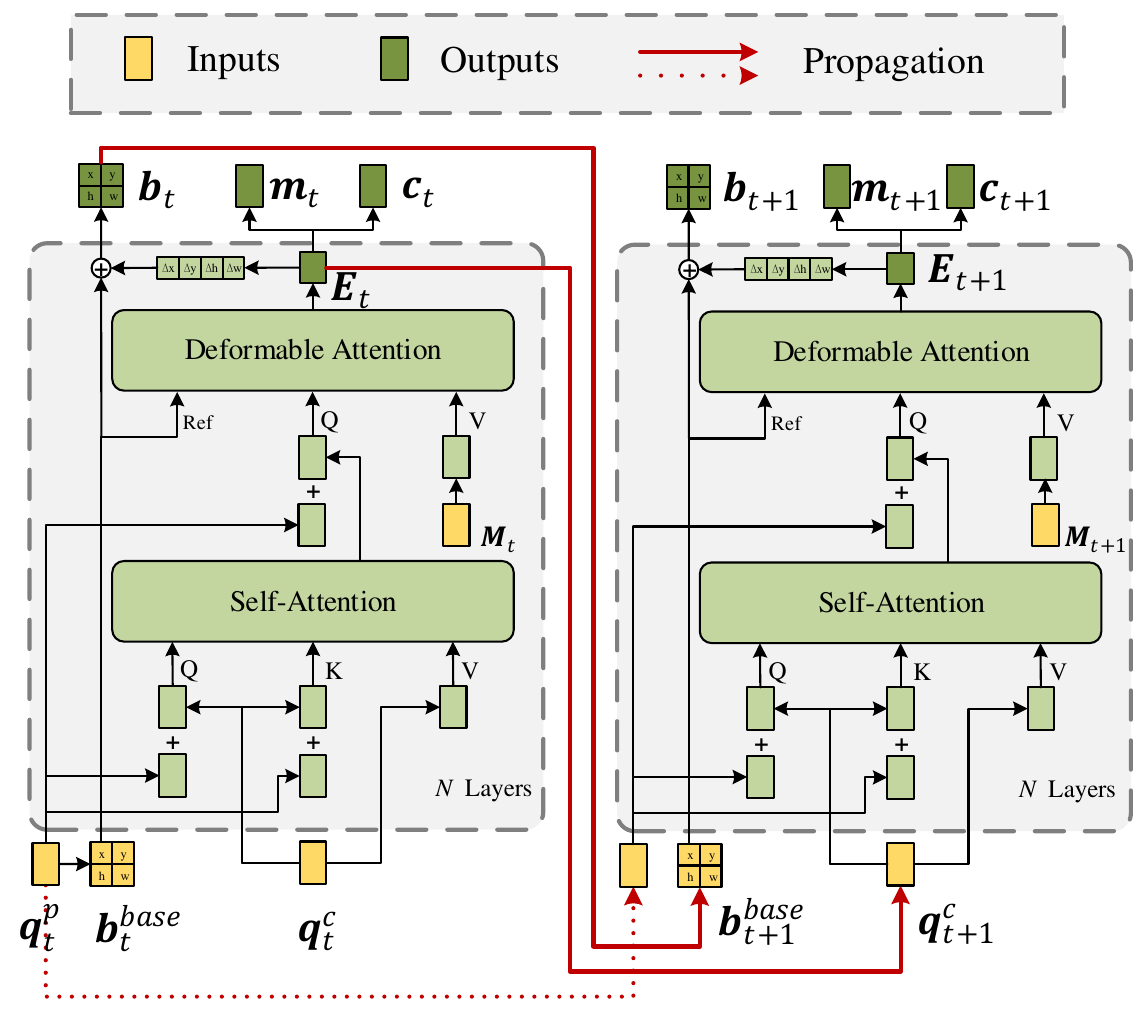}
	\vspace{-10pt}
	\caption{\textbf{Illustration of cross-frame query propagation}. The propagated representations consist of three cues: the output boxes, output embeddings, and position embeddings.}
	\vspace{-3mm}
	\label{fig:query_pro}
\end{figure}

In specific,  given the  $t^{th}$ frame $I_t \! \in \!\mathbb{R}^{3 \times H \times W}$ and its corresponding expression $S$, we separately utilize visual and linguistic backbone to extract their features.
The two features are mapped into the same dimension and fed into an encoder to perform cross-modal fusion using a cross-attention module before encoder layers.
The generated cross-modal features contain critical target awareness, represented by $\bm{M}_t$.
In the decoder, we define two types of queries: content query $\bm{q}_t^c \!\in\! \mathbb{R}^{N_t\times d}$  and position query (\ie, position embedding) $\bm{q}_t^p \!\in\! \mathbb{R}^{N_t\times d}$,  where $N_t$ is the number of queries.
Here, the content query follows the common usage of DETR~\cite{detr}, while the position query is transformed into base values of output boxes, denoted as $\bm{b}^{base}_t\!\in\! \mathbb{R}^{N_t\times 4}$, which \textit{decreases prediction difficulty and benefits model convergence}.
After that, the decoder transforms the queries and cross-modal features into output embedding $\bm{E}_t\!\in\! \mathbb{R}^{N_t\times d}$ (see Fig.~\ref{fig:query_pro} for more details).

On top of the output embedding, a 3-layer feed-forward network (FFN) is used to predict box offset $\bm{b}^{offset}_t\!\in\! \mathbb{R}^{N\times 4}$, which add on the base box coordinate to formulate the final box predictions, \ie, $\bm{b}_t\!=\!\bm{b}^{base}_t\! +\bm{b}^{offset}_t$.
Another 3-layer FFN generates class probabilities $\bm{c}_t\!\in\! \mathbb{R}^{N_t\times c}$, where $c$ is the category number.
For per-frame mask generation, we first employ a cross-modal FPN~\cite{referformer} to perform multi-scale interactions between linguistic features and visual feature maps.
A new FFN then encodes the output embedding into parameters of the mask head, which performs three-layer $1\!\times\! 1$ convolution on the generated FPN feature map, producing mask $\bm{m}_t\!\in \! \mathbb{R}^{N_t\times \frac{H}{4} \times \frac{W}{4}}$.

Since there is only one referent object in the video, we can find the best prediction as positive sample by minimizing the matching cost between predictions and ground truth:
\begin{equation}
\label{eq:match}
\small
\mathcal{L}_{match} =  \lambda_{cls}\mathcal{L}_{cls} + \lambda_{box}\mathcal{L}_{box} + \lambda_{mask}\mathcal{L}_{mask},
\end{equation}
where $\mathcal{L}_{cls}$ is the class-related loss using the focal loss~\cite{lin2017focal}. 
$\mathcal{L}_{box}$ represents the box-related loss that combines $\mathcal{L}_1$ loss and GIoU loss~\cite{rezatofighi2019generalized}.
$\mathcal{L}_{mask}$ is the mask-related loss that sums up DICE loss~\cite{milletari2016v} and binary mask focal loss.
$\lambda_{cls}$, $\lambda_{box}$ and $\lambda_{box}$ are the corresponding loss coefficients.
After completing the matching, we optimize the network using the loss function $\mathcal{L}_{match}$ for positive samples while letting negative samples predict the $\varnothing$ class.

\begin{table*}[t]
	\centering
	\small
	\resizebox{0.98\textwidth}{!}{
		\setlength\tabcolsep{8pt}
		\renewcommand\arraystretch{1.0}
		\begin{tabular}{l||c|c|ccc|ccc}
			\hline\thickhline
             \multirow{2}{*}{Method} &  \multirow{2}{*}{Backbone} & \multirow{2}{*}{Online/Offline} & \multicolumn{3}{c|}{Refer-Youtube-VOS} &\multicolumn{3}{c}{Refer-DAVIS$_{17}$}  \\
			  & & &$\mathcal{J}$\&$\mathcal{F}$& $\mathcal{J}$ & $\mathcal{F}$& $\mathcal{J}$\&$\mathcal{F}$ & $\mathcal{J}$ & $\mathcal{F}$ \\ 
			\hline
              \hline
                CMSA~\cite{CMSA} &ResNet-50& online & 34.9  & 33.3 & 36.5 & 34.7  & 32.1 & 37.2\\
                CMSA + RNN~\cite{CMSA}&ResNet-50& online &36.4 & 34.7 & 38.0  & 40.2 & 36.9 & 43.4 \\
                URVOS~\cite{URVOS}&ResNet-50& online & 47.2 & 45.2 & 49.1  & 51.6&  47.2 & 55.9 \\
                MLSA~\cite{mlsa}&ResNet-50 & \textcolor{mygray}{offline} &49.7 & 48.4 & 50.9 & 57.9 & 53.8 & 62.0 \\
                ReferFormer~\cite{referformer} &ResNet-50& \textcolor{mygray}{offline} & 55.6& 54.8& 56.5&58.5&\textbf{55.8}& 61.3\\
                \rowcolor[gray]{0.9}
                OnlineRefer &ResNet-50& online & \textbf{57.3}	& \textbf{55.6}	& \textbf{58.9} & \textbf{59.3}	& \underline{55.7}	& \textbf{62.9} \\
                \hline
                PMINet + CFBI~\cite{ding2021progressive} & Ensemble & \textcolor{mygray}{offline} & 54.2 & 53.0 & 55.5 & - & - & - \\
                CITD~\cite{liang2021rethinking} & Ensemble & \textcolor{mygray}{offline} & 61.4 & 60.0 & 62.7  & - & - & -\\
                ReferFormer~\cite{referformer}  & Swin-L & \textcolor{mygray}{offline} & 62.4 & 60.8 & 64.0 & 60.5 &57.6 &63.4\\
                \rowcolor[gray]{0.9}
                OnlineRefer & Swin-L & online &  \textbf{63.5}	& \textbf{61.6}	& \textbf{65.5} & \textbf{64.8}	& \textbf{61.6}	& \textbf{67.7} 	\\
                \hline
                MTTR ($\omega$=12)~\cite{mttr} & Video-Swin-T & \textcolor{mygray}{offline} & 55.3 & 54.0 & 56.6 & - & - & - \\
                ReferFormer ($\omega$=5)~\cite{referformer} & Video-Swin-T & \textcolor{mygray}{offline} & 59.4 & 58.0 & 60.9 & - & - & - \\
                ReferFormer ($\omega$=5)~\cite{referformer} & Video-Swin-B & \textcolor{mygray}{offline} & 62.9 & \textbf{61.3} & 64.6 & 61.1 &58.1 &64.1\\
                \rowcolor[gray]{0.9}
                OnlineRefer ($\omega$=2) & Video-Swin-B & semi-online &  \textbf{62.9}& \underline{61.0} &\textbf{64.7}	& \textbf{62.4}	&\textbf{59.1} & \textbf{65.6} \\
			\hline \thickhline
	\end{tabular} }
	\vspace{-8pt}
	\caption{\textbf{The quantitative evaluation on Refer-Youtube-VOS  and Refer-DAVIS$_{17}$}, with region similarity $\mathcal{J}$, boundary accuracy $\mathcal{F}$, and average of $\mathcal{J}$\&$\mathcal{F}$. The best results are in bold and the second ones are underlined.}
	\vspace{-4mm}
	\label{table:sota_ytb}
\end{table*}

\subsection{Cross-frame Query Propagation}
\label{sec:query propagation}

As described above, the decoder of query-based referring segmentation progressively refines the base coordinates from the position query into the final prediction along decoder layers.
Inspired by this, we additionally consider the refinement domain in \textit{temporal axis}, because the target box predicted from the last frame can be a better reference coordinate.
Therefore, we propose the cross-frame query propagation, whose pipeline is illustrated in Fig.~\ref{fig:query_pro}.

Given the outputs of the last frame, we first filter out informative representations that contain rich target awareness.
In specific, we choose the query with the highest class score, and its index is denoted as $\hat{n}$:
\begin{equation}
\small
\vspace{-3pt}
    \hat{n} = \mathop{\arg\max}\limits_{n\in N_t} (\bm{c}_n).
\vspace{-1pt}
\end{equation}
Here, as the first frame follows the original setting~\cite{deformable} and uses $N_1\!=\!5$ learned queries, we can determine one query from multiple ones.
This also leads to the subsequent frames containing only $N_t\!=\!1$ query (\ie, $t\!>\!1
$), which is retained across the entire video.


Once the index is obtained,  we propagate three kinds of corresponding target cues from $t^{th}$ frame to ${(t\!+\!1)}^{th}$  frame, including the prediction box, output embedding, and position embedding.
The box and position embedding represent the explicit position information, which can be transformed as the base coordinate and position query of  ${(t\!+\!1)}^{th}$  frame, which is seamlessly inserted query-based referring segmentation.
The output embedding gathers the semantic information of the target, which is transformed as the content query of  ${(t\!+\!1)}^{th}$  frame. Formally, the propagation process is:
\begin{equation}
\label{eq:propagation}
\small
\begin{aligned}
    \bm{b}_{t+1}^{base} &= \bm{b}_{t, \hat{n}} ~~~~~~~~~~~~~~~~~\in \mathbb{R}^{1\times 4}, \\
    \bm{q}_{t+1}^p &= \bm{q}^p_{t, \hat{n} } ~~~~~~~~~~~~~~~~~\in \mathbb{R}^{1\times d}, \\
    \bm{q}_{t+1}^c &= \mathcal{F}^{FFN}(\bm{E}_{t, \hat{n}}) ~\in \mathbb{R}^{1\times d}, \\
\end{aligned}
\end{equation}
where $\mathcal{F}^{FFN}$ refers to one 3-layer FFN.
Using the query propagation in multiple training frames, the matching cost is independently computed from each frame and the final loss is averaged by the frame number.

\noindent\textbf{Discussion.} 
In a streaming video, it is common for the referent object to enter our view in the middle frames.
In other words, if an object is invisible in the first frame, our model will generate an \textit{empty query} that contains no object for propagation.
Despite this, OnlineRefer still handles well the entrance objects, as shown in the qualitative results of Fig.~\ref{fig:result_ytb}.
This indicates that our cross-frame query propagation aims to provide a good prior while the cross-modal understanding in the query-based referring segmentation still plays an important role in referring prediction.

\subsection{Extension to Semi-Online Model}
\label{sec:extension}

In addition to the frame-by-frame pattern,  we extend OnlineRefer into a more generalized framework that follows a clip-by-clip paradigm.
Its primary motivation is to be capable of large video-based backbones. 
Different from the existing offline methods that independently process each clip, our approach provides query propagation between clips to achieve cross-clip object association.
In this work, we define the new framework as \textit{semi-online} method.

Formally, given the $i^{th}$ video clip $V_i\! \in \!\mathbb{R}^{I \times 3 \times H \times W}$, where $I$ represents the clip length, we feed it into the query-based referring segmentation. 
Our semi-online model first extracts multi-frame visual features using a video-based backbone, \ie, Video Swin Transformer~\cite{videoswin}, and perform cross-modal interaction between visual and linguistic embedding in the encoder.
As the input query of the semi-online model is the same as our online model in the decoder, we then repeat the input query by $I$ times to adapt to the multi-frame referring prediction.
Thus, the semi-online model outputs  multi-frame boxes $\bm{b}_i\!\in\!\mathbb{R}^{I \times N_i \times 4}$,  masks $\bm{m}_i\!\in\!\mathbb{R}^{I \times N_i \times H \times W}$, and classes $\bm{c}_i\!\in\!\mathbb{R}^{I \times N_i \times c}$.
The outputs can be regarded as $N_i$ trajectory predictions on $I$ frames.
Finally, we find the positive sequence from $N_i$ predictions by calculating the matching cost Eq.~\ref{eq:match} and optimize the network.
During the query propagation, we only transfer the high-score query of the last frame into the next clip.

\begin{table*}[t]
	\centering
	\small
	\resizebox{0.98\textwidth}{!}{
		\setlength\tabcolsep{6pt}
		\renewcommand\arraystretch{1.0}
		\begin{tabular}{l||c|c|ccccc|cc}
			\hline\thickhline
			Method& Backbone &Online/Offline& P0.5 & P0.6 & P0.7 & P0.8 & P0.9 & Overall IoU & Mean IoU\\ 
			\hline
            \hline
            Hu \etal~\cite{hu2016segmentation} &VGG-16 &\textcolor{mygray}{offline} &34.8 &23.6 &13.3 &3.3 &0.1 &47.4&35.0 \\
            Gavrilyuk \etal~\cite{gavrilyuk2018actor}& I3D &\textcolor{mygray}{offline} &47.5 &34.7 &21.1 &8.0 &0.2 &53.6 &42.1  	  \\
            CMSA + CFSA~\cite{ye2021referring} &ResNet-101&\textcolor{mygray}{offline} & 48.7& 43.1 & 35.8& 23.1 &5.2& 61.8& 43.2 \\
            ACAN~\cite{wang2019asymmetric} &I3D&\textcolor{mygray}{offline} & 55.7 &  45.9 & 31.9  & 16.0  & 2.0 & 60.1 & 49.0 \\
            CMPC-V~\cite{liu2021cross} &I3D&\textcolor{mygray}{offline} & 65.5 &59.2 &50.6 &34.2 &9.8& 65.3& 57.3\\
            ClawCraneNet~\cite{liang2021clawcranenet}& ResNet-50/101&\textcolor{mygray}{offline} &70.4 &67.7 & 61.7 & 48.9 & 17.1 & 63.1 & 59.9  \\
            MTTR ($\omega$=10)~\cite{mttr} & Video-Swin-T &\textcolor{mygray}{offline} & 75.4 & 71.2 & 63.8 & 48.5 & 16.9 & 72.0 & 64.0 \\
            ReferFormer ($\omega$=5)~\cite{referformer} & Video-Swin-T &\textcolor{mygray}{offline} &  82.8 & 79.2 & 72.3 & 55.3 & 19.3 & 77.6 & 69.6\\
            ReferFormer ($\omega$=5)~\cite{referformer} & Video-Swin-B &\textcolor{mygray}{offline} & 83.1 & 80.4 & 74.1 & 57.9 & 21.2 & 78.6 & 70.3 \\ 
            \rowcolor[gray]{0.9}
            OnlineRefer ($\omega$=5) & Video-Swin-B & semi-online & \textbf{83.1}	& \underline{80.2}	& \underline{73.4}	& \underline{56.8}	&\textbf{21.7}	&\textbf{79.6}	&\textbf{70.5}\\
			\hline \thickhline
	\end{tabular} }
	\vspace{-6pt}
	\caption{\textbf{The quantitative evaluation on A2D-Sentences}, with Precision@K, overall IoU and Mean IoU.}
	\vspace{-2mm}
	\label{table:sota_a2d}
\end{table*}

\begin{table*}[t]
	\centering
	\small
	\resizebox{0.98\textwidth}{!}{
		\setlength\tabcolsep{6pt}
		\renewcommand\arraystretch{1.0}
		\begin{tabular}{l||c|c|ccccc|cc}
			\hline\thickhline
			Method& Backbone & Online/Offline& P0.5 & P0.6 & P0.7 & P0.8 & P0.9 & Overall IoU & Mean IoU\\ 
			\hline
            \hline
            Hu \etal~\cite{hu2016segmentation} &VGG-16 &\textcolor{mygray}{offline} &63.3 &35.0 &8.5 &0.2 &0.0 &54.6 &52.8 \\
            Gavrilyuk \etal~\cite{gavrilyuk2018actor}& I3D &\textcolor{mygray}{offline} &69.9 & 46.0 & 17.3 & 1.4 & 0.0 & 54.1 & 54.2  	  \\
            CMSA + CFSA~\cite{ye2021referring} &ResNet-101&\textcolor{mygray}{offline} & 76.4 & 62.5 & 38.9 & 9.0 & 0.1 & 62.8 & 58.1 \\
            ACAN~\cite{wang2019asymmetric} &I3D&\textcolor{mygray}{offline} & 75.6 & 56.4 & 28.7 & 3.4 & 0.0 & 57.6 & 58.4\\
            CMPC-V~\cite{liu2021cross} &I3D&\textcolor{mygray}{offline} & 81.3 & 65.7 & 37.1 & 7.0 & 0.0   & 61.6 & 61.7\\
            ClawCraneNet~\cite{liang2021clawcranenet}& ResNet-50/101&\textcolor{mygray}{offline} &88.0 & 79.6 &56.6 & 14.7 & 0.2 & 64.4 & 65.6  \\
            MTTR ($\omega$=10)~\cite{mttr} & Video-Swin-T &\textcolor{mygray}{offline} & 93.9 & 85.2 & 61.6 & 16.6 & 0.1 & 70.1 & 69.8 \\
            ReferFormer ($\omega$=5)~\cite{referformer} & Video-Swin-T &\textcolor{mygray}{offline} &  95.8 &  89.3 & 66.8 & 18.9 & 0.2 & 71.9 & 71.0\\
            ReferFormer ($\omega$=5)~\cite{referformer} & Video-Swin-B &\textcolor{mygray}{offline} & 96.2 &90.2 & 70.2 & 21.0 & 0.3 & 73.0 & 71.8 \\ 
            \rowcolor[gray]{0.9}
            OnlineRefer ($\omega$=5) & Video-Swin-B & semi-online & \underline{96.1}	& \textbf{90.4} &	\textbf{71.0}	& \textbf{21.9} &	\underline{0.2}	& \textbf{73.5}	& \textbf{71.9}\\
			\hline \thickhline
	\end{tabular} }
	\vspace{-6pt}
	\caption{\textbf{The quantitative evaluation on JHMDB-Sentences}, with Precision@K, overall IoU and Mean IoU.}
	\vspace{-4mm}
	\label{table:sota_jhmdb}
\end{table*}

\section{Experiment}

\subsection{Dataset and Metric}
\noindent \textbf{Dataset.}
We evaluate our approach on four popular benchmarks: \textbf{Refer-Youtube-VOS}~\cite{URVOS}, \textbf{Refer-DAVIS$_{17}$}~\cite{khoreva2019video}, \textbf{A2D-Sentences}~\cite{gavrilyuk2018actor}, and \textbf{JHMDB-Sentences}~\cite{gavrilyuk2018actor}.
{Refer-Youtube-VOS} expands the large-scale video object segmentation benchmark Youtube-VOS~\cite{xu2018youtube} using texutal desciptions. It consists of 3,975 videos and  27,899 expressions. 
{Refer-DAVIS$_{17}$} extends another video object segmentation benchmark DAVIS$_{17}$~\cite{pont20172017}, which has 90 videos (60 for training and 30 for testing) and more than 1,500 expressions.
A2D-Sentences and JHMDB-Sentences are developed by adding additional textual descriptions on the original action and actor datasets A2D~\cite{xu2015can} and JHMDB~\cite{jhuang2013towards}.
A2D-Sentences includes 3,782 videos and 6,655 expressions, where each video has 3-5 frames annotated with pixel-level segmentation mask.
JHMDB-Sentences contains 928 videos, each being described by a corresponding expression (a total of 928 sentences).

\noindent\textbf{Evaluation Metric.}
On Refer-Youtube-VOS and Refer-DAVIS$_{17}$, we employ region similarity $\mathcal{J}$, contour accuracy $\mathcal{F}$, and their average value $\mathcal{J} \& \mathcal{F}$ as our metrics.
Since  ground-truth annotations of Refer-Youtube-VOS validation are currently inaccessible, our predictions are uploaded to the official server for evaluation.
On A2D-Sentences and JHMDB-Sentences, we  adopt  Overall IoU, Mean IoU, and precision@K to evaluate our method. Overall IoU is the ratio between the total intersection and the total union area over all the test samples. Mean IoU computes the averaged IoU over all the test samples. 
Precision@K measures the percentage of test samples whose IoU scores are higher than a threshold K, where K$\in$[0.5, 0.6, 0.7, 0.8, 0.9].

\subsection{Experiment Details}

\noindent\textbf{Model.}
We implement different visual backbones for feature extraction, such as ResNet~\cite{resnet}, Swin Transformer~\cite{liu2021swin}, Video Swin Transformer~\cite{videoswin}.
RoBERTa~\cite{liu2019roberta} is adopted as the text encoder, while  its parameters are frozen during the entire training stage.
The feature maps of the last three stages are used in the encoder and FPN.
We utilize 4 encoder layers and 4 decoder layers with dimension $d\!=\!256$.
The query number of the first frame is set to $N_1\!=\!5$.
The coefficients for losses are $\lambda_{cls} = 2$, $\lambda_{box}$ = 5, $\lambda_{mask}=2$.

\noindent\textbf{Training.}
AdamW optimizer~\cite{loshchilov2017decoupled} is used to optimize our model with an initial learning rate of $1e\!-\!5$, except for the visual backbone with a learning rate of $5e\!-\!6$.
The training procedure runs for 6 epochs with the learning rate decays divided by 10 at the $3^{th}$ and $5^{th}$ epoch.
The data augmentation techniques  include random horizontal flip, random resize, random crop, and photometric distortion.
 Each frame is resized such that the shorter side at least has a size of 320 and the longer side at most has a size of 576.

For Refer-Youtube-VOS, we randomly sample 3 frames during training online models.
 The inputs of semi-online models are 3 clips, each one containing a window size of 2 (denoted as $\omega\!=\!2$).
In order to improve training stability, we feed 2 frames/clips into the online/semi-online model before the $4^{th}$ epoch.
For Refer-DAVIS$_{17}$, we directly report the results using the model trained on Refer-Youtube-VOS without fine-tuning.
For A2D-Sentences, we use 2 training clips and increase into 3 clips at the $4^{th}$ epoch, where each clip has 5 frames (\ie, $\omega=5$) with the annotated target frame in the middle.
For JHMDB-Sentences, the model trained on A2D-Sentences is directly employed for evaluation without fine-tuning.
For a fair comparison, our model is pre-trained on Ref-COCO~\cite{yu2016modeling}.

\begin{figure}
\centering
\small
	\includegraphics[width=0.9\linewidth]{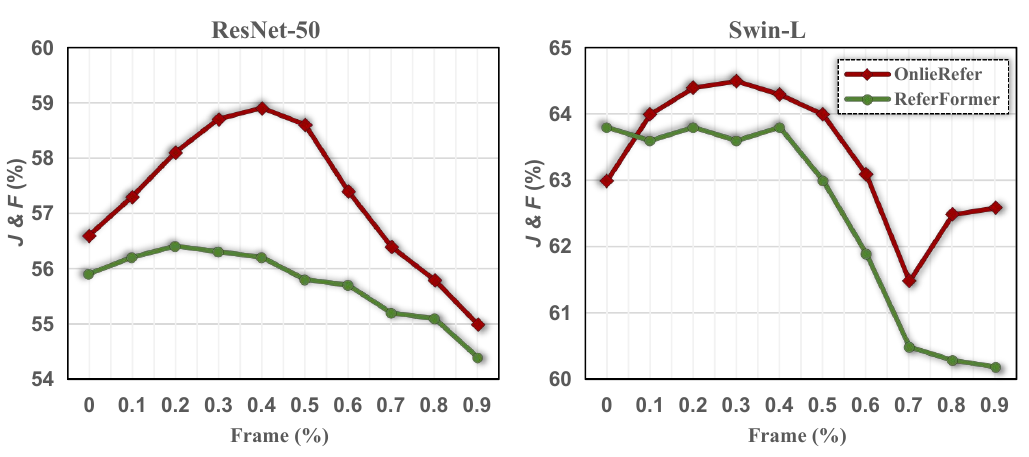}
	\vspace{-12pt}
	\caption{\textbf{Frame-wise J\&F score} on Refer-Youtube-VOS. OnlineRefer has similar J\&F accuracy with ReferFormer at the first frame but obtains better performance in subsequent frames benefiting from a good prior. }
	\vspace{-6 mm}
	\label{fig:score}
\end{figure}

\begin{figure*}
\centering
	\includegraphics[width=0.95\linewidth]{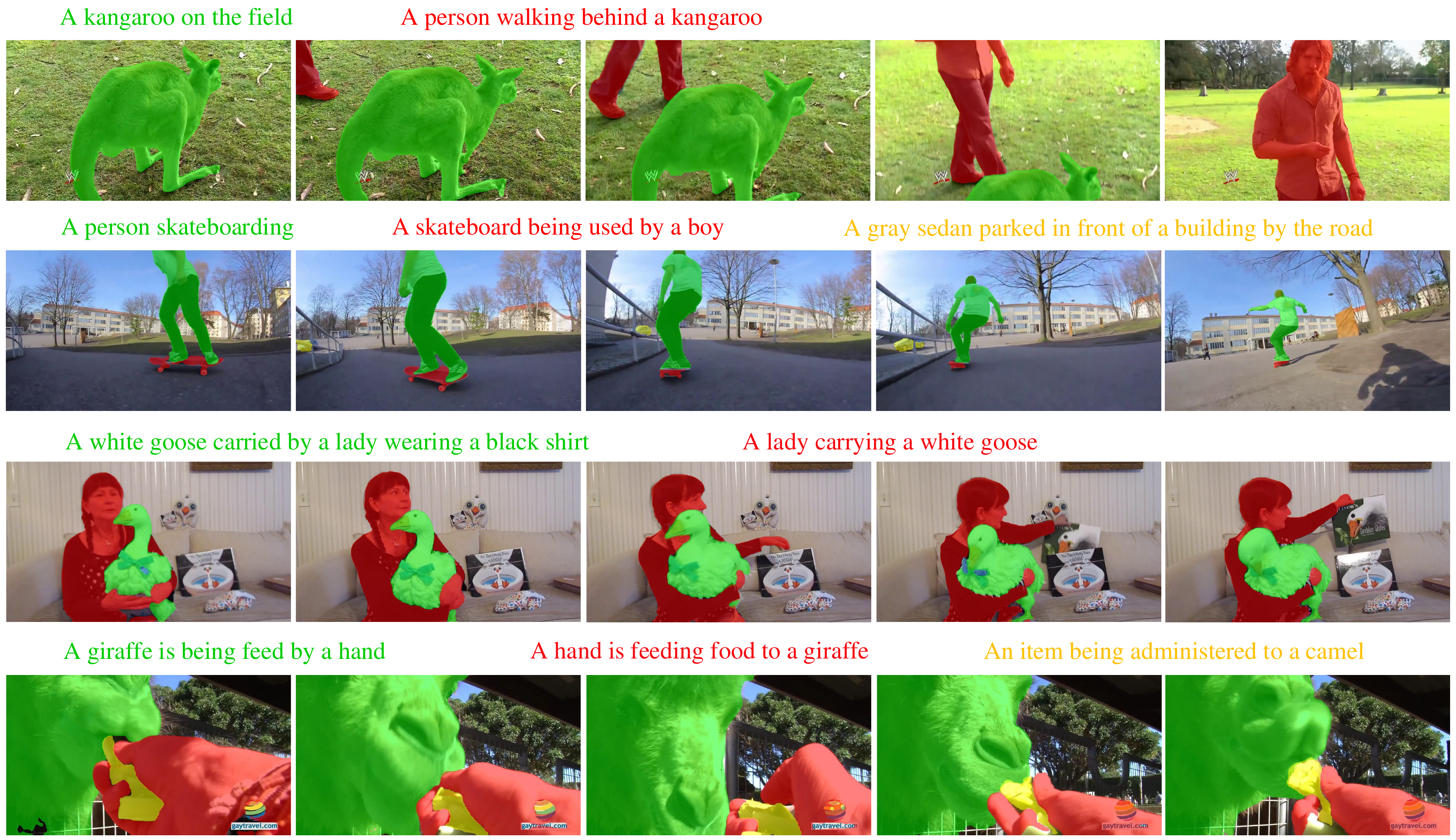}
	\vspace{-10pt}
	\caption{\textbf{Qualitative results from Refer-Youtube-VOS}. OnlineRefer accurately segments the referent object under various challenging scenes, \eg, object occlusion or exit, appearance and size variation, and visually-similar objects.}
	\vspace{-4pt}
	\label{fig:result_ytb}
\end{figure*}

\noindent\textbf{Inference.}
Since there is no gap between training and inference in our method, we directly output the predicted segmentation masks using the well-trained model without post-process.
The prediction is selected with the highest positive score.
In the semi-online paradigm, the clip numbers remain the same with the training setup.
More experiment details are included in supplementary materials.

\subsection{Comparison to State-of-the-art}

\noindent\textbf{Refer-Youtube-VOS \& Refer-DAVIS$_{17}$.}
We compare our method with existing approaches on Refer-Youtube-VOS and Refer-DAVIS$_{17}$, as shown in Table~\ref{table:sota_ytb}.
Note that PMINet~\cite{ding2021progressive} and CITD~\cite{liang2021rethinking} are top-2 solutions in 2021 Refer-Youtube-VOS Challenge.
We can see that the earlier works~\cite{CMSA,URVOS} usually employ an online manner, while the offline methods currently become mainstream due to better performance, such as MLSA~\cite{mlsa}, MTTR~\cite{mttr}, and ReferFormer~\cite{referformer}.
Surprisingly, our online model outperforms all offline methods on two datasets under all metrics.
In specific, on Refer-Youtube-VOS, OnlineRefer with backbone ResNet-50 and Swin-L achieves $\mathcal{J} \& \mathcal{F}$ of 57.3 and 63.5.
They are the highest accuracy among the models using the same backbone.
Notably, our online model performance using Swin-L ($\mathcal{J} \& \mathcal{F}$: \textbf{63.5}) is higher than ReferFormer using Video Swin-B backbone ($\mathcal{J} \& \mathcal{F}$: 62.4).
When the model is directly evaluated on Refer-DAVIS$_{17}$, it achieves the best scores ($\mathcal{J} \& \mathcal{F}$: \textbf{64.8}), which surpasses ReferFormer by a large margin.
Overall, these impressive results sigificiantly demonstrate the effectiveness of the complete-online pipeline in referring video object segmentation.

Furthermore, attaching OnlineRefer on top of a Video Swin-B backbone formulates a semi-online model. The results are displayed in the last row of Table~\ref{table:sota_ytb}, which show that the new semi-online model leads to promising performance on two datasets, especially for the contour accuracy $\mathcal{F}$.
This phenomenon also happens on other online models, which means that the regional prior of propagated boxes benefits producing high-quality segmentation masks.


In addition, we show the frame-wise $\mathcal{J} \& \mathcal{F}$ score curve on Refer-Youtube-VOS using ResNet-50 and Swin-L in Fig.~\ref{fig:score}.
Both OnlineRefer and ReferFormer achieve similar accuracy in the first frame.
But in subsequent frames, our OnlineRefer has a performance gain of around $1\!\sim\! 2$ point.
This clearly approves that the employment of explicit references can improve the referring segmentation performance.

\begin{table}[t]
	\centering
	\small
 \! \! 
	\resizebox{0.48\textwidth}{!}{
		\setlength\tabcolsep{6pt}
		\renewcommand\arraystretch{1.0}
		\begin{tabular}{cc|ccc}
			\hline\thickhline
        Query Update & Position Update &$\mathcal{J}$\&$\mathcal{F}$& $\mathcal{J}$ & $\mathcal{F}$\\
        \hline
 	$\times$      &$\times$       & 49.8 & 49.5	& 50.1 \\
        \rowcolor[gray]{0.9}
        $\checkmark$    & $\times$   & \textbf{57.3} & \textbf{55.8}  & \textbf{58.8} \\
        $\checkmark$    &$\checkmark$& 32.1 & 29.7	& 34.6 \\
			\hline 
	\end{tabular} }
	\vspace{-6pt}
	\caption{\textbf{The update strategy of query and position} on Refer-Youtube-VOS, in terms of $\mathcal{J}$,  $\mathcal{F}$.
 }
	\vspace{-4mm}
	\label{table:aba_update}
\end{table}

 \begin{table*}[t!]
 \small
    \begin{center}
    \setlength{\tabcolsep}{4pt}
    \begin{subtable}[t]{0.3\linewidth}
        \centering
        \begin{tabular}{p{2.6cm}|ccc}
        \hline\thickhline
        Propagation Method &$\mathcal{J}$\&$\mathcal{F}$& $\mathcal{J}$ & $\mathcal{F}$\\
        \hline
        \textit{w/o} Propagation & 49.3 & 47.4	& 51.2 \\
 	  Concatenation & 55.2 & 53.8 & 56.7 \\
        Fixed		  &55.0  & 53.6	& 56.5 \\
        \rowcolor[gray]{0.9}
         Ours         & \textbf{57.3} & \textbf{55.8} & \textbf{58.8} \\
        \hline
        \end{tabular}
        \vspace{-1mm}
        \caption{Comparison on propagation method. }
        \label{table:3a}
    \end{subtable}
    \  \  \ \ \ \ 
    \setlength{\tabcolsep}{4pt}
    \begin{subtable}[t]{0.3\linewidth}
        \centering
        \begin{tabular}{c|ccc}
        \hline\thickhline
         Propagation Number &$\mathcal{J}$\&$\mathcal{F}$& $\mathcal{J}$ & $\mathcal{F}$ \\
        \hline
 	  Top-4 & 55.2 & 53.7 & 56.7 \\
        Top-3 & 55.9 & 54.4 & 57.4 \\
        Top-2 & 56.3 & 54.9 & 57.7 \\
        \rowcolor[gray]{0.9}
        Top-1 & \textbf{57.3} & \textbf{55.8} & \textbf{58.8} \\
        \hline
        \end{tabular}
        \vspace{-1mm}
        \caption{Comparison on propagation number. }
        \label{table:3b}
    \end{subtable}
     \  \  \    \  \
    \setlength{\tabcolsep}{4pt}
    \begin{subtable}[t]{0.3\linewidth}
        \centering
        \begin{tabular}{c|ccc}
        \hline\thickhline
        Initial Queries &$\mathcal{J}$\&$\mathcal{F}$& $\mathcal{J}$ & $\mathcal{F}$\\
        \hline
 	1 & 54.1 & 52.6 & 55.6 \\
        3 & 55.5 & 54.0 & 56.9 \\
        \rowcolor[gray]{0.9}
        5 & \textbf{57.3} & \textbf{55.8} & \textbf{58.8} \\
        8 & 56.9 & 55.2 & 58.5 \\
        \hline
        \end{tabular}
        \vspace{-0.5mm}
        \caption{Comparison on sampler frames. }
        \label{table:3c}
    \end{subtable}
\end{center}
\vspace{-18pt}
  \caption{
    \textbf{Ablation studies of different propagation designs} on Refer-Youtube-VOS, in terms of region similarity $\mathcal{J}$, boundary accuracy $\mathcal{F}$, and average of $\mathcal{J}$\&$\mathcal{F}$. The best results are in bold.
    }
    \vspace{-4mm}
\end{table*}

\noindent\textbf{A2D-Sentences \& JHMDB-Sentences.}
We further present comparisons on the A2D-Sentences benchmark in Table~\ref{table:sota_a2d}.
As the dataset is only annotated keyframes, existing methods generally follow an offline paradigm, which process clip-wise referring prediction without any cross-clip association.
In contrast, our OnlineRefer is able to link all video clips by query propagation, \ie, the semi-online framework.
With the backbone Video Swin-B and window size of $\omega\!=\!5$, our semi-online model obviously exceeds all offline methods over IoU metrics and keeps enough competitiveness over precision metrics as well.

The semi-online model is directly evaluated on JHMDB-Sentences without finetuning to  demonstrate the generality of our method. In Table~\ref{table:sota_jhmdb}, OnlineRefer achieves competitive performance compared to all other offline methods.
In specific, OnlineRefer leads to higher IoU scores but comparable precision scores.
Considering the clip gap on A2D-Sentences and JHMDB-Sentences, the above results demonstrate the potential of semi-online model.


\subsection{Qualitative Results}

In Fig.~\ref{fig:result_ytb}, we show several typical referring segmentation results of OnlineRefer from Refer-Youtube-VOS.
The first video sequence is more challenging because the two referring objects become occluded and invisible in some frames.
Taking the walking person as an example, it requires our online model to avoid the predicted empty box in the first frame causing the target missing in subsequent frames.
Despite the difficulty, our OnlineRefer successfully segments out the target with sharp boundaries, showing strong correctness ability.
In other scenes, the referring objects also face various challenges, such as appearance variation, pose deformation, and visually-similar objects.
Otherwise, our OnlineRefer performs well in these difficult scenarios.

\subsection{Ablation Study}
\label{sec:ablation}

To offer a deep insight into our OnlineRefer, we conduct ablation studies to analyze the effectiveness of each component.
If not specialized, we report the online model performance on Refer-Youtube-VOS using ResNet-50.

\noindent\textbf{Importance of query updating.}
To investigate the effect of updating strategy in Eq.~\ref{eq:propagation}, we perform experiments whether updating query and position embedding in Table~\ref{table:aba_update}.
From the first row, discarding  both query and position updating only  achieves 49.8 over $\mathcal{J} \& \mathcal{F}$.
After that, adding the query update achieves remarkable performance improvement (+6.7 on $\mathcal{J} \& \mathcal{F}$), and reaches the best score.
However, updating position embedding largely hinders model performance (-25.2 on $\mathcal{J} \& \mathcal{F}$), as shown in the last row of  Table~\ref{table:aba_update}.
The problem demonstrates that fixed position embedding plays an important role in cross-frame object association.

\noindent\textbf{Different propagation methods.}
We then analyze different query propagation methods in Table~\ref{table:3a}.
Removing propagation (\ie, \textit{w/o} propagation) leads to a frame-based referring segmentation pipeline, which results in a large performance drop (-6.0 on $\mathcal{J} \& \mathcal{F}$).
`Concatenation' represents concatenating the one propagated query and the initial query set, like MOTR~\cite{motr} and TrackFormer~\cite{trackformer}.
`Fixed' refers to giving up the query selection and keeping the initial number of queries, like InsPro~\cite{inspro}.
Both two have a slight performance decrease.
These results approve the effectiveness of our simple and heuristic-free
query propagation.

\noindent\textbf{Number of propagation query.}
It is also of interest to explore the impact of different propagation query numbers.
As shown in Table~\ref{table:3b}, we vary the query number from top-4 to top-1, where top-1 is our setting in OnlineRefer.
Note that propagating top-5 queries equals the `fixed' method in Table~\ref{table:3a}.
It is obvious that with the top query number decreasing, the performance on $\mathcal{J} \& \mathcal{F}$ is gradually improved.
Overall, applying one query on cross-frame propagation is enough and significant  for the inter-frame association.

\noindent\textbf{Effect of initial queries.}
OnlineRefer starts with a set of initial queries in the first frame, which is further propagated across the entire video.
To study its influence, we use a relatively small number of queries, as shown in Table~\ref{table:3c}.
We can see that fewer queries bring fewer proposals, further leading to lower $\mathcal{J} \& \mathcal{F}$ scores.
However, the performance of more queries becomes flattened after $N\!=\!5$.
Empirically, in this work, we set the initial query number as $N\!=\!5$.

\section{Conclusion}


In this paper, we proposed a simple, elegant, and strong baseline for online referring video object segmentation, named OnlineRefer.
It broke up the widely accepted tradition that only offline models can handle well the challenging referring understanding task.
OnlineRefer includes two crucial parts, query-based referring segmentation, and query propagation.
The query-based referring segmentation outputs box, mask, and category based on input queries, while query generation part updates the output set as new queries.
By iteratively using two parts, the objects in all video frames are automatically associated and predicted.
To be compatible with video-based backbones, we developed a semi-online model that associates and predicts referent object clip by clip.
The experiments are conducted on Refer-Youtube-VOS, Refer-DAVIS$_{17}$, A2D-Sentences and JHMDB-Sentences and our OnlineRefer  shows the state-of-the-art performance on the four benchmarks.

{\small
\bibliographystyle{ieee_fullname}
\bibliography{egbib}
}

\end{document}